%% file: main.tex
\documentclass{article}

\usepackage{microtype}
\usepackage{graphicx}
\usepackage{subfigure}
\usepackage{booktabs} 

\usepackage[all]{xy}

\usepackage{hyperref}

\usepackage[accepted]{icml2020}

\input{macros}

\begin{document}

\twocolumn[
\icmltitle{Inverse Reinforcement Learning via Matching of Optimality Profiles}

\icmlsetsymbol{equal}{*}

\begin{icmlauthorlist}
\icmlauthor{Luis Haug}{ethz}
\icmlauthor{Ivan Ovinnikov}{ethz}
\icmlauthor{Eugene Bykovets}{ethz}
\end{icmlauthorlist}

\icmlaffiliation{ethz}{Department of Computer Science, ETH Zürich, Switzerland}

\icmlcorrespondingauthor{Luis Haug}{lhaug@inf.ethz.ch}

\icmlkeywords{Machine Learning, Reinforcement Learning, Inverse
    Reinforcement Learning, Deep Reinforcement Learning, Deep
    Learning}

\vskip 0.3in
]

\printAffiliationsAndNotice{}  

\begin{abstract}
    The goal of inverse reinforcement learning (IRL) is to infer a
    reward function that explains the behavior of an agent performing
    a task. The assumption that most approaches make is that the
    demonstrated behavior is near-optimal. In many real-world
    scenarios, however, examples of truly optimal behavior are scarce,
    and it is desirable to effectively leverage sets of demonstrations
    of suboptimal or heterogeneous performance, which are easier to
    obtain. We propose an algorithm that learns a reward function from
    such demonstrations together with a weak supervision signal in the
    form of a distribution over rewards collected during the
    demonstrations (or, more generally, a distribution over cumulative
    discounted future rewards). We view such distributions, which we
    also refer to as optimality profiles, as summaries of the degree
    of optimality of the demonstrations that may, for example, reflect
    the opinion of a human expert. Given an optimality profile and a
    small amount of additional supervision, our algorithm fits a
    reward function, modeled as a neural network, by essentially
    minimizing the Wasserstein distance between the corresponding
    induced distribution and the optimality profile. We show that our
    method is capable of learning reward functions such that policies
    trained to optimize them outperform the demonstrations used for
    fitting the reward functions.
\end{abstract}

\input{intro}

\input{related_work}

\input{setting}
\input{problem}

\input{method}

\input{algorithm}
\input{experiments}

\input{conclusions}

\bibliography{references}
\bibliographystyle{icml2020}

\onecolumn
\appendix
\input{appendix}

\end{document}

%% file: macros.tex
\usepackage[utf8]{inputenc} 
\usepackage[T1]{fontenc}    
\usepackage{hyperref}       
\usepackage{url}            
\usepackage{booktabs}       
\usepackage{amsfonts}       
\usepackage{nicefrac}       
\usepackage{microtype}

\usepackage{amsmath}
\usepackage{amssymb}
\usepackage{amsthm}

\makeatletter

\newcommand{\Rmnum}[1]{\expandafter\@slowromancap\romannumeral #1@}
\makeatother

\DeclareMathOperator*{\argmin}{arg\,min}

\newcommand{\R}{\mathbb{R}}

\def \argmin {\mathop{\rm arg\,min}}

\newcommand{\cA}{{\mathcal{A}}}

\newcommand{\cD}{{\mathcal{D}}}

\newcommand{\cI}{{\mathcal{I}}}
\newcommand{\cL}{{\mathcal{L}}}
\newcommand{\cM}{{\mathcal{M}}}
\newcommand{\cN}{{\mathcal{N}}}

\newcommand{\cS}{{\mathcal{S}}}
\newcommand{\cR}{{\mathcal{R}}}
\newcommand{\cT}{{\mathcal{T}}}

\newcommand{\cX}{{\mathcal{X}}}
\newcommand{\cY}{{\mathcal{Y}}}

\newcommand{\ba}{{\mathbf{a}}}

\newcommand{\bs}{{\mathbf{s}}}

\newcommand{\by}{{\mathbf{y}}}

%% file: intro.tex
\section{Introduction}
Reinforcement learning has achieved remarkable success in learning
complex behavior in tasks in which there is a clear reward signal
according to which one can differentiate between more or less
successful behavior. However, in many situations in which one might
want to apply reinforcement learning, it is impossible to specify such
a reward function. This is especially the case when small nuances are
important for judging the quality of a certain behavior. For example,
the performance of a surgeon cannot be judged solely based on binary
outcomes of surgeries performed, but requires a careful assessment
e.g.\ of psychomotoric abilities that manifest themselves in subtle
characteristics of motion.

When a reward function cannot be specified, an alternative route is to
use inverse reinforcement learning (IRL) to \emph{learn} it from a set
of demonstrations performing the task of interest. The learned reward
function can then serve as a compact machine-level representation of
expert knowledge whose usefulness extends beyond the possibility to
optimize a policy based on it. For example, the primary purpose of
learning a reward function might be to use it for performance
assessment in educational settings.

The assumption of most existing IRL approaches is that the
demonstrated behavior is provided by an expert and near-optimal (for
an unknown ground truth reward function), and the ensuing paradigm is that
the reward function to be found should make the demonstrations look
optimal.

In fact, one can argue that it may not even be desirable to learn a
reward function solely based on optimal behavior, especially when the
final goal is not primarily to optimize a policy that imitates expert
behavior, but rather to be able to accurately assess performance
across a wide range of abilities. Moreover, it is often unrealistic to
get access to many optimal demonstrations in real-world settings,
while access to demonstrations of heterogeneous quality, potentially
not containing any examples of truly optimal behavior at all, may be
abundant. It is then natural that the role of the expert shifts from a
provider of demonstrations to a provider of an assessment of the
quality of demonstrations performed by non-experts.

In this paper, we investigate the possibility that such an expert
assessment is provided in the form of a distribution over rewards seen
during the demonstrations, which we call an \emph{optimality
    profile}. The basic, and slightly naive, idea is simply that good
policies tend to spend much time in regions with high reward, while
bad policies tend to spend much time in regions with low reward, which
is reflected in the optimality profile. As this naive view may not be accurate in
settings with sparse or delayed rewards, we consider, more generally,
distributions over cumulative discounted future rewards, which create
an association between states and the long-term rewards they lead to
eventually.

The IRL paradigm that we propose is then to find a reward function
subject to the requirement that it is consistent with a given optimality
profile, in the sense that the induced distribution over reward values
seen in the demonstrated trajectories approximates the optimality
profile.

We think of optimality profiles as succinct summaries of expert
opinions about the degree of optimality of demonstrations. We believe
that the elicitation of such optimality profiles from experts is much
more feasible than direct labeling of a large number of state-action
pairs with reward values (based upon which one could use methods more
akin to traditional regression for reward function learning). However,
we point out that our method still needs a certain amount of additional
supervision to cut down the degrees of freedom that remain when the
requirement of matching an optimality profile is satisfied. Our
experiments indicate that our method is robust to a certain amount of
noise in the optimality profile.

%% file: related_work.tex
\section{Related work}
\label{sec:related-work}

\paragraph{Inverse reinforcement learning} 
Inverse reinforcement learning, introduced by \cite{ng2000algorithms}
\cite{abbeel2004apprenticeship} \cite{ziebart2008maximum} and expanded
to high-dimensional tasks in
\cite{wulfmeier2015maximum},\cite{finn2016guided},
\cite{ho2016generative}, \cite{fu2017learning}, aims to learn a reward
function based on near-optimal task demonstrations. The optimality of
the demonstrations usually required to train valid reward functions is
often problematic due to the high cost of obtaining such
demonstrations. Hence, a number of algorithms exploiting a lower
degree of optimality have been proposed.

\paragraph{Suboptimal demonstrations} 
A limited number of works have explored the setting of learning a
reward function from suboptimal trajectories. An extreme case of
learning from suboptimal demonstrations is presented in
\cite{shiarlis_inverse_2016}.  The proposed method makes use of
maximum entropy IRL to match feature counts of successful and failed
demonstrations for a reward which is linear in input features. A
different approach is taken in \cite{levine_reinforcement_2018} which
introduces a probabilistic graphical model for trajectories enhanced
with a binary optimality variable which encodes whether state-action
pairs are optimal.  This approach reconciles the maximum causal
entropy IRL \cite{ziebart2013principle} method and the solution of the
forward RL problem by considering the binary optimality variable
dependent on the reward function to describe the suboptimal
trajectories. The authors of \cite{brown_trex_2019} and subsequently
\cite{brown_drex_2019} propose a method which uses an objective based
on pairwise comparisons between trajectories in order to induce a
ranking on a set of suboptimal trajectories. They demonstrate the
extrapolation capacity of the model by outperforming the trajectories
provided at training time on the ground truth episodic returns. While
they are able to learn high quality reward functions, they still
require a substantial amount of supervision in the form of pairwise
comparisons.

\paragraph{Preference based learning}
The ranking approach presented in \cite{brown_trex_2019} is an
instance of a larger class of preference based learning methods.  Due
to the scarcity of high-quality demonstrations required for using IRL,
several methods have proposed learning policies directly based on
supervision that encodes preferences. For example,
\cite{akrour2011preference}, \cite{wilson2012bayesian}, \cite{warnell2018deep}
work in a learning scenario where the expert provides a set of
non-numerical preferences between states, actions or entire
trajectories \cite{wirth2017survey}. The use of pairwise trajectory ranking for the purpose
of learning a policy that plays Atari games has been shown in 
\cite{christiano_deep_2017}. The method relies on a large number 
of labels provided by the annotator in the policy learning stage.
The successor of this method, \cite{ibarz_reward_2018}, 
provides a solution to this problem by using a 
combination of imitation learning and preference based learning.

\paragraph{Optimal transport in RL}
The use of the Wasserstein metric between value distributions has been
explored in \cite{bellemare2017distributional}, which takes a
distributional perspective on RL in order to achieve improved results
on the Atari benchmarks. The authors of \cite{xiao2019wasserstein}
adapt the dual formulation of the optimal transport problem similar to
a Wasserstein GAN in for the purposes of imitation learning.

%% file: setting.tex
\section{Setting}
\label{sec:setting}
We consider an environment modelled by a \emph{Markov decision
    process} $\cM = (\cS, \cA, T, P_0, R)$, where $\cS$ is the state
space, $\cA$ is the action space, $T$ is the family of transition distributions on
$\cS$ indexed by $\cS \times \cA$ with $T_{s, a}(s')$ describing the
probability of transitioning to state $s'$ when taking action $a$ in
state $s$, $P_0$ is the initial state distribution, and
$R: \cS \to \R$ is the reward function. For simplicity, we assume that
$R$ is a function of states only (it could also be defined on
state-action pairs, for example). A \emph{policy} $\pi$ is a map
taking states $s \in \cS$ to distributions $\pi(\cdot | s)$ over
actions, with $\pi(a|s)$ being the probability of taking action $a$ in
state $s$.

\paragraph{Terminology and notation} Given sets $\cX, \cY$, a
distribution $P$ on $\cX$ and a map $f: \cX \to \cY$, we denote by
$f_*P$ the \emph{push-forward} of $P$ to $\cY$ via $f$, i.e., the
distribution given by $(f_*P)(S) = P(f^{-1}(S))$ for $S \subset \cY$.

\subsection{Distributions on state and trajectory spaces}
\label{sec:distr-traj}

In this subsection, we define distributions $\rho_{\cS}, \rho_{\cT}$
on states and trajectory spaces that we will use for the definition of
the reward distributions we are interested in. (These distributions
are quite ``natural'', so the reader not interested in the precise
details may want to jump directly to the next subsection.)

A \emph{trajectory} is a sequence of states
$\bs = (s_0, s_1, \dots) \in \cS^{\infty}$. An MDP $\cM$ and a policy
$\pi$ together induce a distribution on the set of trajectories
$\cS^{\infty}$ as follows:
\begin{equation}
    \label{eq:traj_distr}
    \begin{aligned}
        P_{\mathrm{traj}}(\bs) = \sum_{\ba} P_0(s_0) \cdot \prod_{t \geq 0}
        \pi(a_t|s_t) \cdot P(s_{t + 1} | a_t, s_t),
    \end{aligned}
\end{equation}
where the first sum is over all possible action sequences
$\ba = (a_0, a_1, \dots) \in \cA^{\infty}$. While these trajectories
are a priori infinite, we will be interested in trajectories of length
at most $T$ for some fixed finite horizon $T > 0$, and thus we
consider $\bs, \bs' \in \cS^\infty$ as equivalent if $s_t = s'_t$ for
all $t \leq T$. We also assume that $\cS$ contains a (possibly empty)
subset $\cS_{\mathrm{term}}$ of \emph{terminal} states for which
$T_{s,a}(s) = 1$ for all $a \in \cA$ (meaning that once a trajectory
reaches such a state $s$, it stays there forever), and we consider
$\bs, \bs'$ as equivalent if they agree up until the first occurrence
of a terminal state (in particular, this needs to happen at the same
timestep in both trajectories). These two identifications give rise to
a map
\begin{equation}
    \label{eq:cut_off_map}
    \cS^\infty \to \cT := \bigcup_{t = 0}^{T} \cS^{\times
        (t + 1)}
\end{equation}
which cuts off trajectories after time step $T$ or after the first
occurrence of a terminal state, whichever happens first;
$\cS^{\times k}$ denotes the $k$-fold Cartesian product of $\cS$. We
denote by $\ell(\bs)$ the length of a finite trajectory $\bs$.

By pushing forward the distribution $P_{\mathrm{traj}}$ on
$\cS^\infty$ \eqref{eq:traj_distr} via the cut-off map
\eqref{eq:cut_off_map}, we obtain a distribution on $\cT$, which we
denote by $P_\mathrm{traj}'$. Using that, we finally define a
distribution $\rho_{\cT_1}$ on $\cT \times [0, T]$ by setting
\begin{equation}
    \label{eq:traj_timestep_distr}
    \rho_{\cT_1} (\bs, t) = \frac{P_\mathrm{traj}'(\bs)}{\sum_{\bs' \in \cT}\ell(\bs') P_\mathrm{traj}'(\bs')}
\end{equation}
for every $(\bs, t)$ satisfying $t \leq \ell(\bs) - 1$, and
$P(\bs, t) = 0$ otherwise; it is easy to check that this defines a
distribution. In words, $\rho_{\cT_1}(\bs, t)$ describes the probability of
obtaining $(\bs, t)$ if one first samples a trajectory from $\cT$
according to a version of $P_{\mathrm{traj}}'$ in which probabilities
are rescaled proportionally to trajectory length, and then a time step
$t$ uniformly at random from $[0, \dots \ell(\bs) - 1]$ (see
additional explanations in the supplementary material). By definition,
the distribution \eqref{eq:traj_timestep_distr} has its support
contained in the subset
\begin{equation}
    \cT_1 = \{(\bs, t) \in \cT \times [0, T] ~|~ t \leq \ell(\bs) -1\}
\end{equation}
of $\cT \times [0, T]$, the ``set of trajectories with a marked time
step'', and therefore we view \eqref{eq:traj_timestep_distr} as a
distribution on $\cT_1$. Note that there are natural maps
\begin{equation}
    \label{eq:map_marked_timestep}
    \begin{aligned}
        \Pi_\cS: \cT_1 \to \cS, \quad &(\bs, t) \mapsto s_t\\
        \Pi_\cT: \cT_1 \to \cT, \quad &(\bs, t) \mapsto (s_t, \dots,
        s_{\ell(\bs) - 1})
    \end{aligned}
\end{equation}
taking $(\bs, t)$ to the state at the marked time step $t$, resp.\ the
future of that state from time step $t$ onward. Together with
$\rho_{\cT_1}$, these maps induce distributions
\begin{equation}
    \label{eq:state_occupancy_future_measure}
    \rho_\cS = (\Pi_\cS)_* \rho_{\cT_1}, \qquad \rho_{\cT} = (\Pi_\cT)_* \rho_{\cT_1}
\end{equation}
on $\cS$ and on $\cT$ which we call the \emph{state occupancy measure}
resp.\ the \emph{future measure}, as $\rho_\cT(\bs')$ is the
probability that, if we sample $(\bs, t) \sim \rho_{\cT_1}$, the
future $\Pi_{\mathrm{\cT}}(\bs, t)$ of $\bs$ from time step $t$ onward
is $\bs'$.

\subsection{Distributions induced by the reward function}
\label{sec:distr-induc-reward}
Given the state occupancy measure $\rho_\cS$ on $\cS$, we can view the
reward function $R: \cS \to \R$ as a random variable with distribution
\begin{equation}
    \label{eq:reward_distr}
    P_R = R_* \rho_{\cS}.
\end{equation}
Note that $P_R$ also depends on $\pi$ (because $\rho_\cS$ does), which
we suppress in the notation for simplicity.

The reward function also gives rise to a natural family of
\emph{return} functions $R^{(\gamma)}: \cT \to \R$ indexed by a
discount parameter $\gamma \in [0, 1]$ and taking a trajectory
$\bs \in \cT$ to
$R^{(\gamma)}(\bs) = \sum_{t = 0}^{\ell(\bs) - 1} \gamma^{t}
R(s_{t})$.
In words, the return $R^{(\gamma)}(\bs)$ is the sum of discounted
rewards collected in $\bs$. Together with the distribution $\rho_\cT$
on $\cT$ defined above, the maps $R^{(\gamma)}$ give rise to
\emph{return distributions} on $\R$,
\begin{equation}
    \label{eq:disc_reward_distr}
    P_R^{(\gamma)} = (R^{(\gamma)})_* \rho_\cT
\end{equation}
For $\gamma = 0$, this recovers the reward distribution defined above,
i.e., $P_R^{(0)} = P_R$; see the supplementary material for additional
explanations.

%% file: problem.tex
\section{Problem formulation}
\label{sec:problem-formulation}
The most common formulation of the IRL problem is: ``Find a reward
function $R$ for which the policy generating the given demonstrations
are optimal,'' where optimality of a policy means that it maximizes
the expected return. It is well known that this is an ill-posed
problem, as a policy may be optimal for many reward
functions. Therefore, additional structure or constraints must be
imposed to obtain a meaningful version of the IRL problem. Such
constraints can come in the form of a restriction to functions of low
complexity, such as linear functions, or in the form of an additional
supervision signal such as pairwise comparisons between demonstrations
\citep{brown_trex_2019}.

Our interpretation of the IRL problem is: ``Find a reward function $R$
which is compatible with a supervision signal that provides
information about the degree of optimality of the given
demonstrations.'' We will focus on a specific type of supervision
signal: We assume that, in addition to a set of demonstrations, we
have access to an estimate of the distribution $P_R = R_*\rho_{\cS}$
over ground truth rewards $R(s_t)$ collected when rolling out the
policy that generated the demonstrations. This distribution reflects
how much time the policy spends in regions of the state space with
high or low reward. Although we use the term ``ground truth'', the
idea is that $P_R$ may for example reflect the subjective opinion of
an expert who judges the demonstrations.

\begin{figure}[t]
    \includegraphics[width=\linewidth]{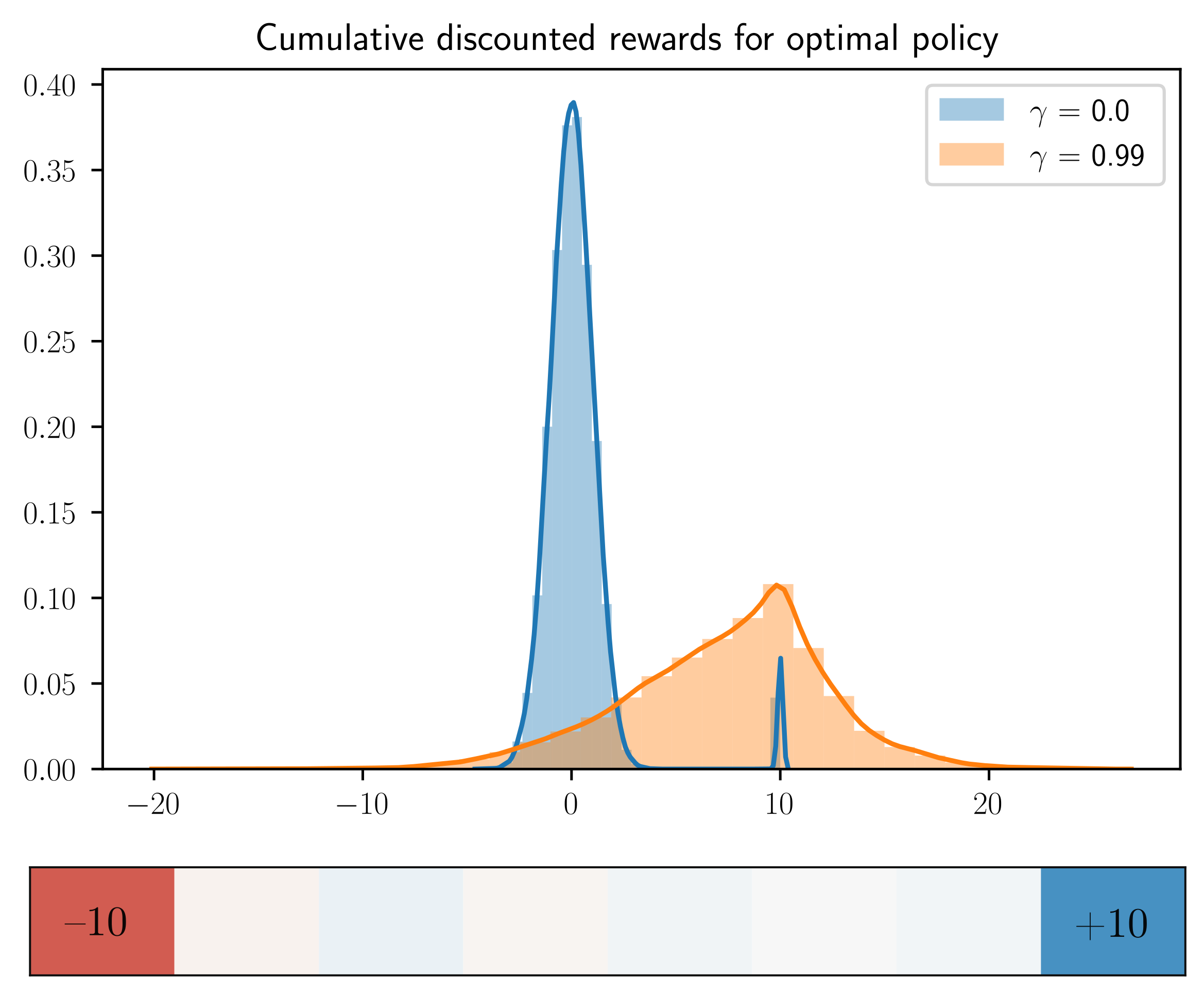}
    \caption{This example illustrates why consideration of
        the distribution of cumulative discounted future rewards may
        be more meaningful than of the reward distribution itself: The
        grid world shown contains a single terminal goal state
        $s_{\mathrm{goal}}$ with $R(s_{\mathrm{goal}}) = +10$, a
        single terminal fail state $s_{\mathrm{fail}}$ with
        $R(s_{\mathrm{fail}}) = - 10$, while for all other states
        reward values are $R(s) \approx 0$. If the set of non-terminal
        states
        $\cS \smallsetminus \{s_{\mathrm{goal}}, s_{\mathrm{fail}}\}$
        is large, both optimal policies, which approach
        $s_{\mathrm{goal}}$ as fast as possible, and highly suboptimal
        policies, which approach $s_{\mathrm{fail}}$ as fast as
        possible, spend most time in this region and their reward
        distributions $P_R$ look very much alike: Most of the mass
        concentrates around 0, and only a small spike at +10 resp.\ -10
        distinguishes them. In contrast, for $\gamma$ tending to 1, the
        distributions $P_R^{(\gamma)}$ for optimal and highly
        suboptimal policies look increasingly different.}
    \label{fig:1d-gridworld}
\end{figure}

A natural generalization is to assume access to the \emph{return}
distribution $P_R^{(\gamma)} = R^{(\gamma)}_* \rho_{\cT}$ for some
$\gamma > 0$. Intuitively, such distributions encode how much time a
policy spends in states from which it will eventually reach a state
with high reward. Values of $\gamma$ close to 0 corresponds to experts
who only take immediate rewards into account, while values $\gamma$
close to 1 corresponds to experts who associate states with rewards
collected in the far future. Figure \ref{fig:1d-gridworld} provides a
toy example illustrating this point.

%% file: method.tex
\section{Method}
\label{sec:method}
The idea on which our method is based is to infer a reward function
from demonstrated trajectories together with an estimate of the
distribution of (cumulative discounted) rewards that is provided as an
input to the algorithm we propose. We refer to this as an
\emph{optimality profile} and denote it by $P_\mathrm{tgt}$ (the
target we want to match). Our algorithm then tries to find a reward
function $R_{\widehat \theta}$ within a space of functions
$\cR_{\Theta} = \{R_{\theta} ~|~ \theta \in \Theta\}$, such as neural networks
of a specific architecture, that minimizes a measure of discrepancy
between the optimality profile and the distribution of (discounted)
rewards which $R_{\widehat \theta}$ induces together with the
empirical distribution on $\cT$:
\begin{equation}
    \label{eq:disc_reward_matching}
    \theta^*  \in \argmin_{\theta \in \Theta}
    \Delta (P_{R_\theta}^{(\gamma)}, P_{\mathrm{tgt}}),
\end{equation}
where $\Delta$ is a metric or divergence measure on the set of
probability distributions on $\R$. In our experiments, we use the
Wasserstein distance $\Delta = W_p$ resp.\ an entropy-regularized
version of it (typically with $p = 2$). We will denote the
corresponding objective by
\begin{equation}
    \label{eq:ot_objective}
    \cL_{\mathrm{ot}} = W_p \big(P_{R_\theta}^{(\gamma)}, P_{\mathrm{tgt}}\big),
\end{equation}
where ``ot'' stands for ``optimal transport''.

\subsection{Symmetry under measure-preserving maps}
\label{sec:symm-under-meas}
It may be useful to think of the optimality profile as being induced
by a ground truth reward function. The idea is that the requirement of
being consistent with the optimality profile induces a strong
constraint on the reward function to be found. Ideally, one would hope
that a function whose induced distribution is close to the optimality
profile should be similar to the ground truth reward function.

However, minima, or approximate minima, of objectives as in
\eqref{eq:disc_reward_matching}, which depend on a function only
through the push-forward of a given measure, are generally far from
unique, due to a large group of symmetries under which such objectives
are invariant. In other words, two functions for which these
push-forwards are close may still be very different.

Indeed, assume for the moment that we could optimize
\eqref{eq:disc_reward_matching} over the entire space of $\R$-valued
functions on $\cS$, and let $\widehat R$ be such a function (we
consider the case $\gamma = 0$ for simplicity). Let now $\Phi$ be an
element of the group of measure-preserving transformations of
$(\cS, \rho_\cS)$, i.e., a map $\Phi: \cS \to \cS$ satisfying
$\Phi_* \rho_\cS = \rho_\cS$. Then we have
\begin{equation}
    \label{eq:invariance_of_push_fwd}
    (\widehat R \circ \Phi)_* \rho_\cS = \widehat R_*
    (\Phi_* \rho_\cS) = \widehat R_* \rho_\cS,
\end{equation}
and hence any objective on functions which only depends on the
function through the push-forward of a given measure is invariant
under the action of this group. In particular,
\begin{equation}
    \label{eq:invariance_of_objective}
    \Delta(\widehat R_* \rho_\cS, P_\mathrm{tgt}) = \Delta((\widehat R \circ \Phi)_*
    \rho_\cS, P_{\mathrm{tgt}})
\end{equation}
for any $P_{\mathrm{tgt}}$. Since the group of measure-preserving maps
is often huge (infinite-dimensional if $\cS$ is a continuous space),
\eqref{eq:invariance_of_objective} implies that for every local
minimum there is a huge subspace of potentially very different
functions with the same value of the objective. (Apart from that, our
objective is also invariant under the action of the group of
measure-preserving maps $(\R, P_{\mathrm{tgt}})$, whose elements
$\phi$ act by post-composition, i.e.,
$\widehat R \mapsto \phi \circ \widehat R$.)

Of course, in reality we do not optimize over the space of \emph{all}
functions $\cS \to \R$, but only over a subspace $\cR_\Theta$
parametrized by a finite-dimensional parameter $\theta$, which is
usually \emph{not} preserved by the action of the group of
measure-preserving maps of $(\cS, \rho)$. However, if $\cR_\Theta$ is
sufficiently complex, ``approximate level sets'' of minima will
typically still be large. We can therefore not expect our objective
\eqref{eq:disc_reward_matching} to pick a function that is
particularly good for our purposes without some additional
supervision.

\subsection{Additional supervision}
\label{sec:addit-superv}
In order to deal with the problem described in Section
\ref{sec:symm-under-meas} in our setting, we propose to use a small
amount of additional supervision in order to break the symmetries.

Specifically, we assume that we have access to a
small set of \emph{pairs} of trajectories
$\{(\bs_1, \bs_1'), \dots, (\bs_m, \bs_m')\} \subset \cT
\times \cT$ which are ordered according to ground truth
function value, i.e., $R^{(\gamma)}(\bs_i) \leq R^{(\gamma)}(\bs_i')$
for all $i$. The corresponding loss component is
\begin{equation}
    \label{eq:pw_loss}
    \cL_{\mathrm{pw}}(R_\theta) = -\sum_{i=1}^m \log
    \frac{\exp\big(R_\theta^{(\gamma)}(\bs_i')\big)}
    {\exp\big(R_\theta^{(\gamma)}(\bs_i)\big)
        + \exp\big(R_\theta^{(\gamma)}(\bs_i')\big)}
\end{equation}
which gives a penalty depending on the extent to which the pairs
$(R^{(\gamma)}_\theta(\bs_i), R^{(\gamma)}_\theta(\bs_i'))$ violate
the ground truth ordering. This type of loss function, which is
related to the classical Bradley--Terry--Luce model of pairwise
preferences \cite{bradley_rank_1952, luce_individual_1959}, has also
been used e.g.\ by \cite{brown_trex_2019,
    christiano_deep_2017,ibarz_reward_2018} for learning from
preferences in (I)RL settings.

Alternatively, or additionally, we assume that we have access to a small set
$\{\bs_1, \dots, \bs_n\} \subset \cT$ of trajectories for which we
have ground truth information about returns
$y_i = R^{(\gamma)}(\bs_i)$, or noisy versions
thereof. Correspondingly, we consider the loss function given by
\begin{equation}
    \label{eq:fix_loss}
    \cL_{\mathrm{fix}}(R_\theta) = \big \Vert \big(R_\theta^{(\gamma)}(\bs_1), \dots,
    R_\theta^{(\gamma)}(\bs_n)\big) - \by \big\Vert_2,
\end{equation}
where $\by = (y_1, \dots, y_n)$.

To summarize, the objective that we propose to optimize is
\begin{equation}
    \label{eq:total_objective}
    \cL_{\mathrm{tot}}(R_\theta) = c_{\mathrm{ot}}
    \cL_{\mathrm{ot}}(R_\theta) + c_{\mathrm{pw}}
    \cL_{\mathrm{pw}}(R_\theta) +
    c_{\mathrm{fix}} \cL_{\mathrm{fix}}(R_\theta),
\end{equation}
a weighted sum of the loss functions discussed above.

%% file: algorithm.tex
\section{Algorithm}
\label{sec:algorithm}
We assume that we are given a set of demonstrations
$\cD = \{\bs_1, \dots, \bs_N\} \subset \cT$ which are sampled from
some policy $\pi$, as well as an optimality profile $P_{\mathrm{tgt}}$
that will serve as the target return distribution for some $\gamma$,
and that reflects for example the opinion of a human expert about the
degree of optimality of the demonstrations in $\cD$.

We will also consider the augmented set of trajectories
\begin{equation}
    \label{eq:augmented_trajs}
    \widetilde{\cD} = \{\bs_1, \bs_1^{[1:]}, \bs_1^{[2:]}, \dots,
    \bs_N^{[\ell_N-2:]}, \bs_N^{[\ell_N - 1:]}\}
\end{equation}
where $\ell_i = \ell(\bs_i)$ and
$\bs^{[j:]} = (s_{j}, \dots, s_{\ell(\bs) - 1})$ is the restriction of
$\bs$ to the time interval $[t, \ell(\bs) - 1]$. In words,
$\widetilde{\cD}$ consists of all trajectories that one obtains by
sampling a trajectory from $\cD$ and a time step
$t \in [0, \ell(\bs)-1]$ and then keeping only futures of that time
step in $\bs$; in particular, we have $\cD \subset
\widetilde{\cD}$. (One could call $\widetilde{\cD}$ the set of ``ends
of trajectories in $\cD$''.)  If $\cD$ is sampled from the trajectory
distribution $P_\mathrm{traj}$ defined in \eqref{eq:traj_distr}, then
$\widetilde{\cD} \subset \cT$ is a sample from the distribution
$\rho_{\cT}$ defined in \eqref{eq:state_occupancy_future_measure}.

For notational simplicity, we will recycle notation and denote the
elements of $\widetilde{\cD}$ again by $\bs_i$, that is,
$\widetilde{\cD} = \{\bs_1, \dots, \bs_M\}$.

\subsection{Stochastic optimization of $\cL_{\mathrm{ot}}(R_\theta)$}
\label{sec:stoch-optimization}
In every round of Algorithm \ref{alg:distr_matching}, we need to
compute an estimate of the optimal transport loss
$\cL_{\mathrm{ot}}(R_\theta)$ for the current function $R_\theta$
based on a minibatch $B \subset \widetilde{\cD}$. Algorithm \ref{alg:ot_loss}
describes how this is done: We first compute the cumulative sum of
discounted future rewards $R^{(\gamma)}_\theta(\bs_i)$ for every
trajectory $\bs_i \in B$, from which we then compute a histogram
$\widehat P_{R_\theta}^{(\gamma)}(B)$ approximating
$P_{R_\theta}^{(\gamma)}$, the return distribution for $R_{\theta}$,
see \eqref{eq:disc_reward_distr}.

In the next step, we compute an approximately optimal transport plan
$G \in \Gamma(\widehat P_{R_\theta}^{(\gamma)}, P_{\mathrm{tgt}})$,
i.e., a distribution $G$ on $\R \times \R$ whose marginals are
$\widehat P_{R_\theta}^{(\gamma)}$ and $P_{\mathrm{tgt}}$ and which
approximately minimizes the entropy-regularized optimal transport
objective
\begin{equation}
    \label{eq:wasserstein_obj}
    \left( \int_{\R \times \R} \vert x-y \vert^p dG'(x, y) \right)^{1/p} -
    \lambda H(G')
\end{equation}
among all distributions
$G' \in \Gamma(\widehat P_{R_\theta}^{(\gamma)},
P_{\mathrm{tgt}})$. Minimizing only the first term in
\eqref{eq:wasserstein_obj} would mean computing an honest optimal
transport plan; the entropy term $\lambda H(G')$ regularizes the
objective, which increases numerical stability. Note that the
distributions for which we need to compute a transport plan live in
dimension 1, which makes this problem feasible
\cite{cuturi_sinkhorn_2013, peyre_computational_2018}. In our
experiments, we use the Python optimal transport package
\cite{flamary2017pot} to compute transport plans.

Once we have the transport plan $G$, we sample an element
$y_{\mathrm{tgt}, j} \sim G(R_{\theta}^{(\gamma)}(\bs_{i_j}), \cdot)$
for every trajectory $\bs_{i_j} \in B$; this is a distribution on $\R$
which encodes where the probability mass that
$\widehat P_{R_\theta}^{(\gamma)}$ places on
$R_{\theta}^{(\gamma)}(\bs_{i_j}) \in \R$ should be transported
according to the transport plan $G$. Finally, we use the
$y_{\mathrm{tgt}, j}$ to estimate $\cL_{\mathrm{ot}}(R_\theta; B)$.

In spirit, our algorithm has similarity with Wasserstein GANs
\cite{arjovsky_wasserstein_2017}, which also attempt to match
distributions by minimizing an estimate of the Wasserstein distance
(with $p = 1$). In contrast to them, we do not make use of
Kantorovich-Rubinstein duality to fit a discriminator (which worked
much worse for us), but rather work directly with 1d optimal transport
plans.

\begin{algorithm}[tb]
    \caption{Estimation of the optimal transport loss $\cL_{\mathrm{ot}}$}
    \label{alg:ot_loss}
    \begin{algorithmic}[1]
        \REQUIRE
        $B = \{\bs_{i_1}, \dots, \bs_{i_b}\} \subset \widetilde{\cD},
        P_\textrm{tgt}$ \STATE Compute
        $y_j = R_\theta^{(\gamma)}(\bs_{i_j})$ for $j = 1, \dots, b$
        \STATE Use the $y_i$ to get an estimate $\widehat P_{R_\theta}^{(\gamma)}(B)$ of
        $P_{R_\theta}^{(\gamma)}$
        \STATE Compute a transport plan $G \in \Gamma\big(\widehat
        P_{R_\theta}^{(\gamma)}(B), P_\mathrm{tgt}\big)$
        \STATE Sample
        $y_{\mathrm{tgt}, j} \sim G\big(R_{\theta}^{(\gamma)}(\bs_{i_j}\big),
        \cdot)$ for $j = 1, \dots, b$
        \STATE $\cL_{\mathrm{ot}}(R_\theta; B) = \Big(\sum_{j=1}^b
            \big\vert R_{\theta}^{(\gamma)}(\bs_{i_j}) - y_{\mathrm{tgt}, j}\big\vert^p\Big)^{1/p}$
    \end{algorithmic}
\end{algorithm}

\begin{algorithm}[tb]
    \caption{Matching of optimality profiles}
    \label{alg:distr_matching}
    \begin{algorithmic}[1]
        \REQUIRE $\cD$, $P_{\mathrm{tgt}}$, $\cI_{\mathrm{pw}}$, $\cY_{\mathrm{fix}}$
        \STATE Initialize weights $\theta$
        \FOR{$i=1, \dots, n_{\mathrm{epochs}}$}
        \STATE Sample a minibatch $B \subset \widetilde{\cD}$
        \STATE Compute $\cL_{\mathrm{ot}}(R_\theta; B)$,$\cL_{\mathrm{pw}}(R_\theta; \cI_{\mathrm{pw}})$,\,$\cL_{\mathrm{fix}}(R_\theta; \cY_{\mathrm{fix}})$
        \STATE Compute $\nabla_\theta \cL_{\mathrm{tot}}(R_\theta)$,
        see \eqref{eq:total_objective}
        \STATE Update weights: $\theta \leftarrow \theta - \eta
        \nabla_\theta \cL_{\mathrm{tot}}(R_\theta)$
        \ENDFOR
    \end{algorithmic}
\end{algorithm}

\subsection{Distribution matching with additional supervision}
\label{eq:distr-matching}
The full distribution matching procedure is summarized by Algorithm
\ref{alg:distr_matching}. In addition to the demonstrations $\cD$ and
the optimality profile $P_\mathrm{tgt}$, it takes as its input (a) a
set $\cY_\mathrm{fix} = \{y_{i_j}, \dots, y_{i_n} \in \R\}$ of labels
for a subset $\cD_\mathrm{fix} \subset \cD$, and/or (b) a set
$\cI_{\mathrm{pw}} = \{(j_1, j_1'), \dots, (j_m, j_m') \in [1,
M]^{\times 2}\}$ of ordered pairs of indices of elements of $\cD$
with respect to which
$\cL_{\mathrm{fix}}(R_\theta; \cY_{\mathrm{fix}})$ \eqref{eq:fix_loss}
resp.\ $\cL_\mathrm{pw}(R_\theta; \cI_\mathrm{pw})$ \eqref{eq:pw_loss}
and their gradients are computed in every optimization epoch. The
reason we require these additional sourced of supervision is to break
symmetries caused by measure-preserving transformations.
   
In every epoch, we estimate the gradient of the total loss
\eqref{eq:total_objective} with respect to $\theta$ and then take a
step in the direction of the negative gradient. Note that when
estimating the gradient $\nabla_\theta \cL_{\mathrm{ot}}(R_\theta)$
with respect to $\theta$, we treat the target values
$y_{\mathrm{tgt}, j}$ as constant (line 5 of Algorithm
\ref{alg:ot_loss}), although they really depend on
$R_\theta$ as well through the optimal transport plan that was
computed using an estimate of $P_{R_\theta}^{(\gamma)}$.

%% file: experiments.tex
\section{Experiments}
\label{sec:experiments}

\begin{figure*}[t]
    \begin{subfigure}[]{
            \includegraphics[width=0.32\linewidth]{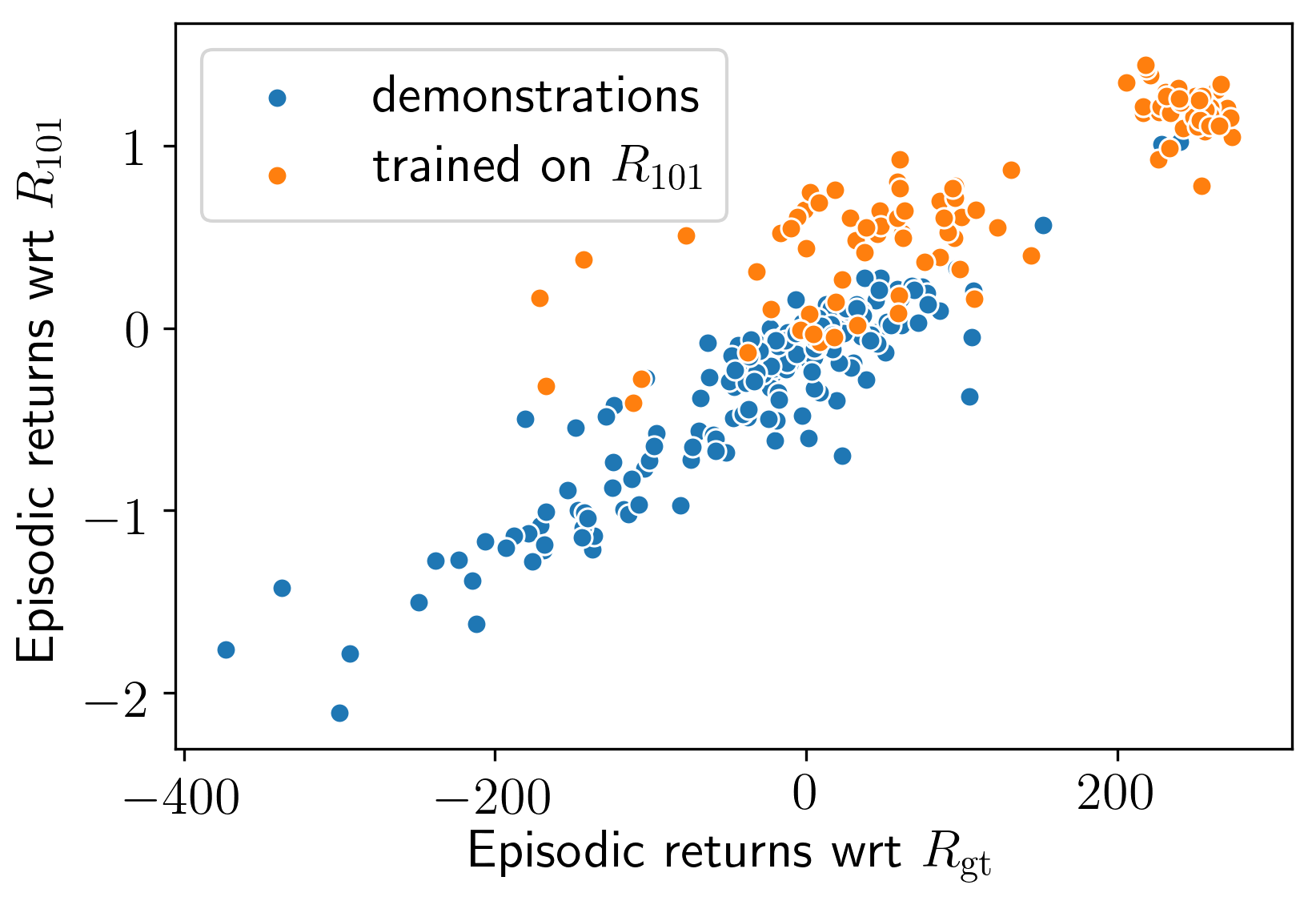}
            \label{fig:104400_evaluation_demos}
        }
    \end{subfigure}
    \begin{subfigure}[]{
            \includegraphics[width=0.32\linewidth]{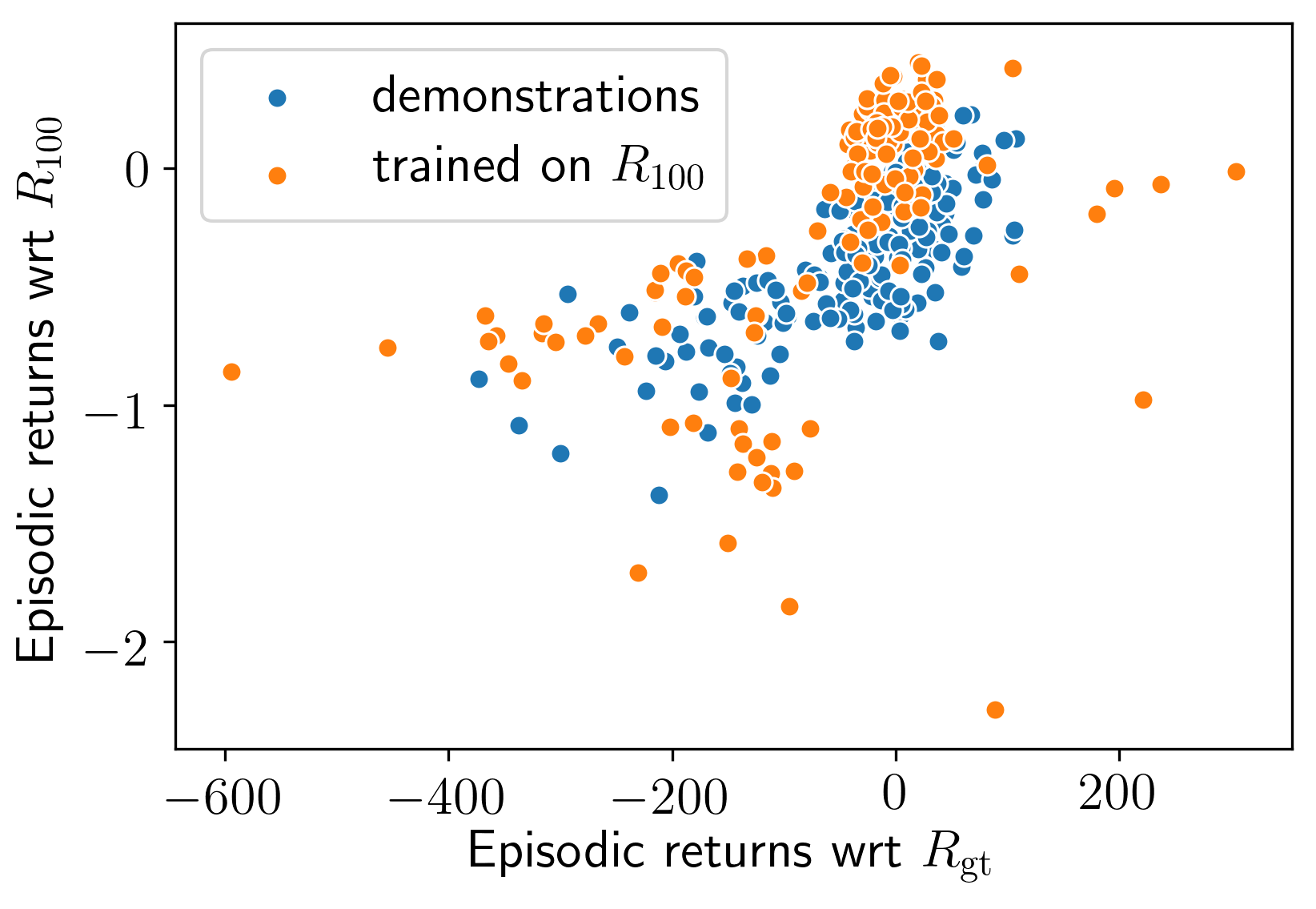}
            \label{fig:200248_evaluation_demos}
        }
    \end{subfigure}
    \begin{subfigure}[]{
            \includegraphics[width=0.32\linewidth]{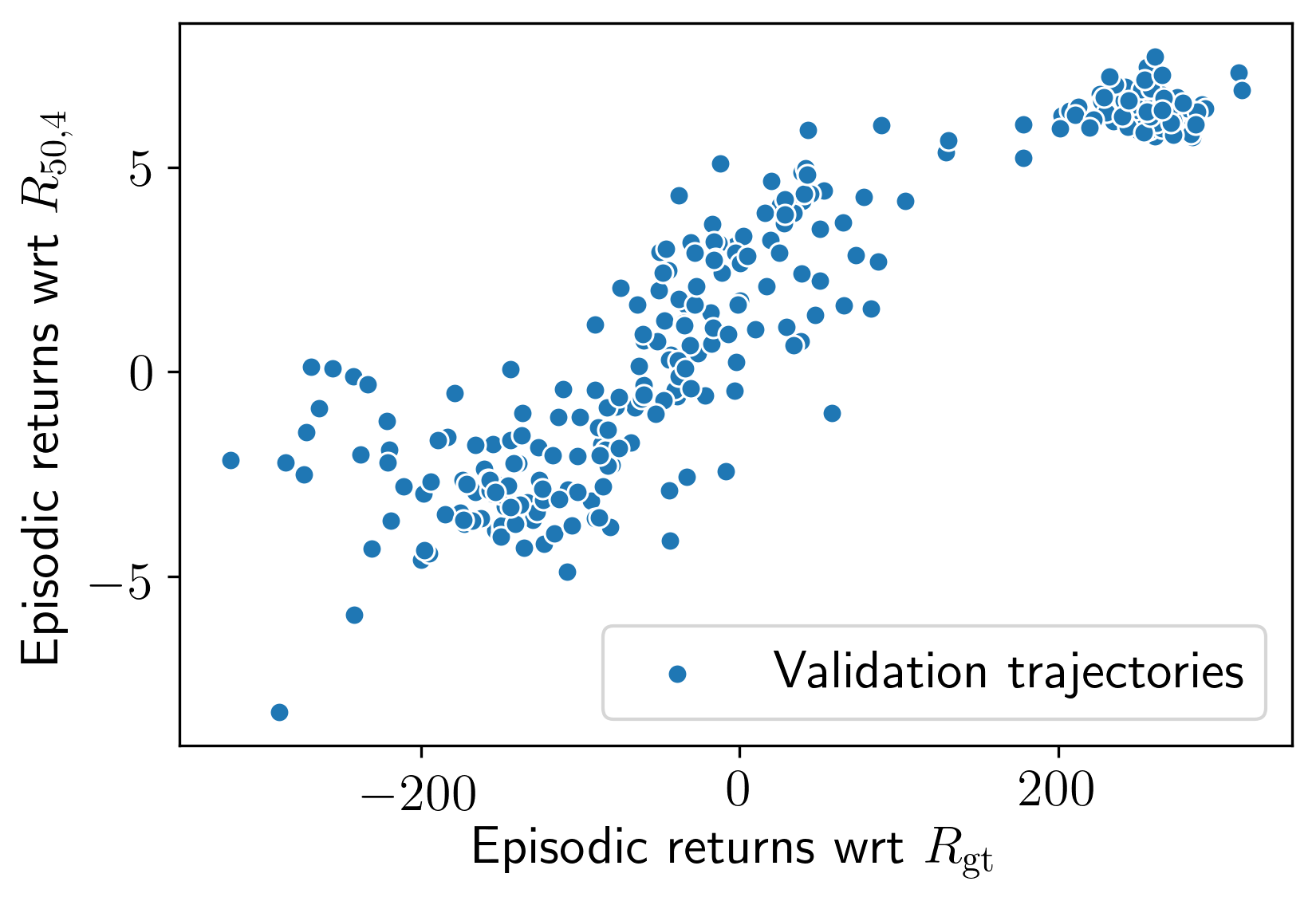}
            \label{fig:162815_validation}
        }
    \end{subfigure}
    \begin{subfigure}[]{
            \includegraphics[width=0.32\linewidth]{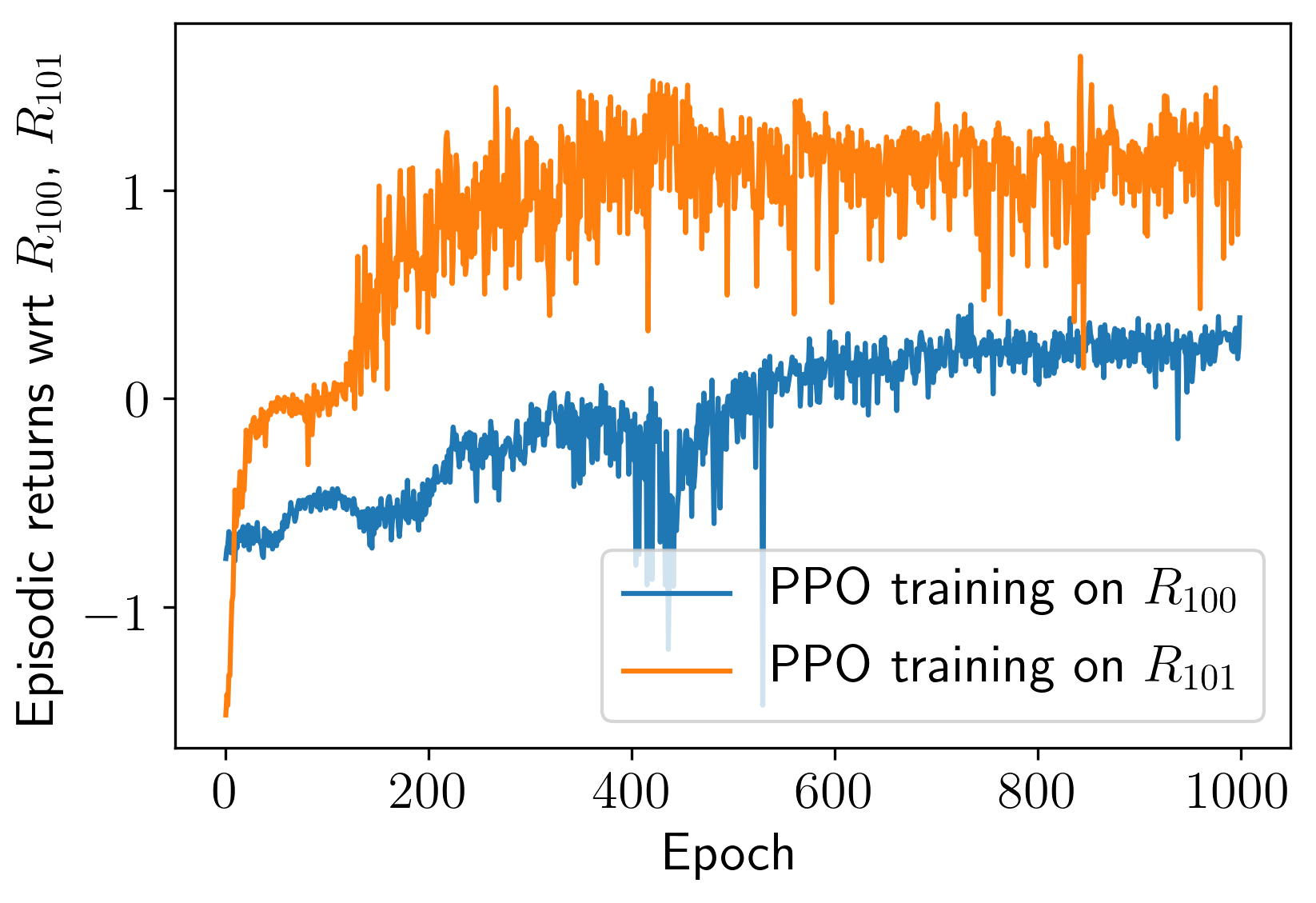}
            \label{fig:104400_200248_training_returns}
        }
    \end{subfigure}
    \begin{subfigure}[]{
            \includegraphics[width=0.32\linewidth]{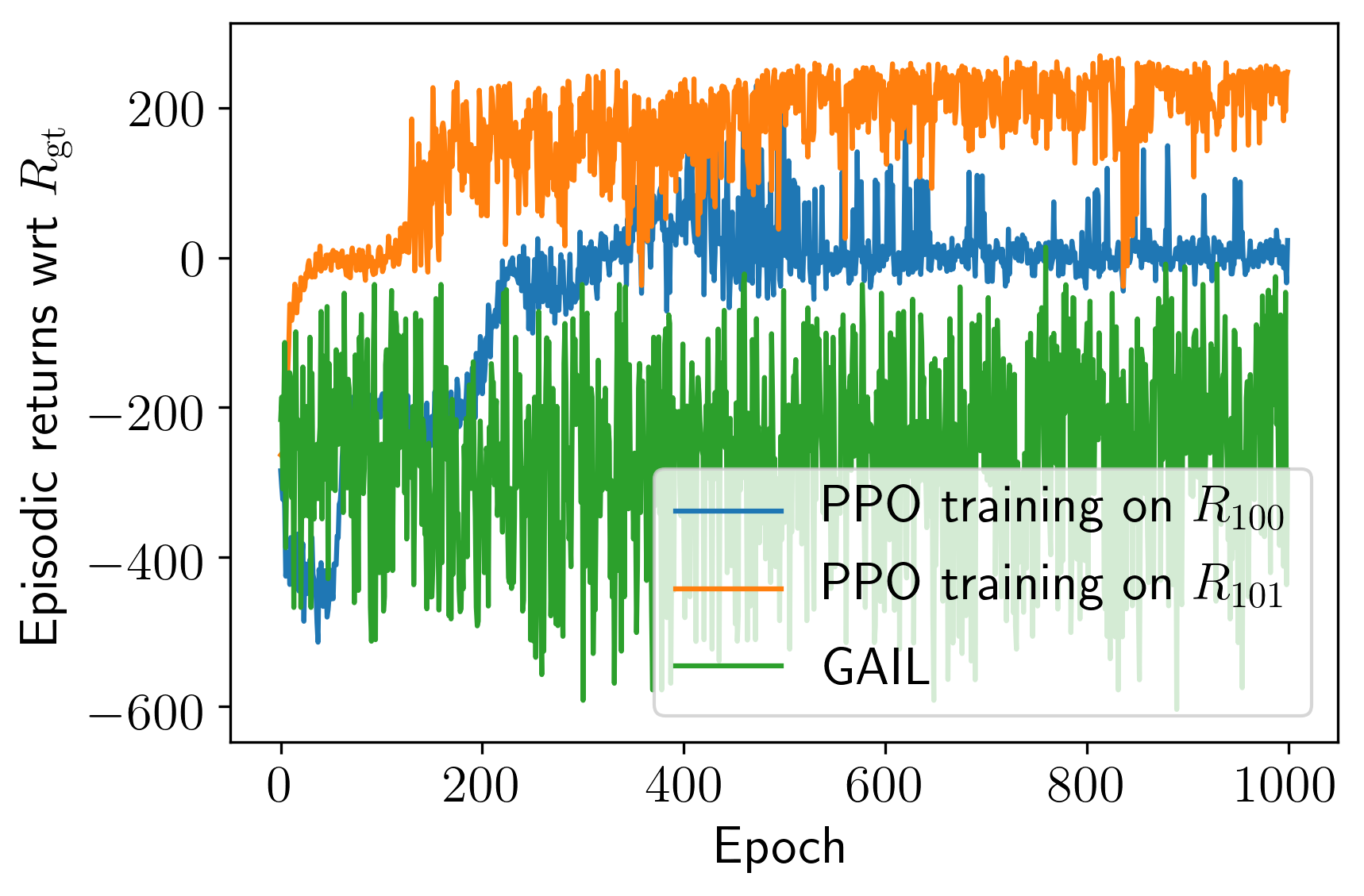}
            \label{fig:104400_200248_training_returns_gt}
        }
    \end{subfigure}
    \begin{subfigure}[]{
            \includegraphics[width=0.32\linewidth]{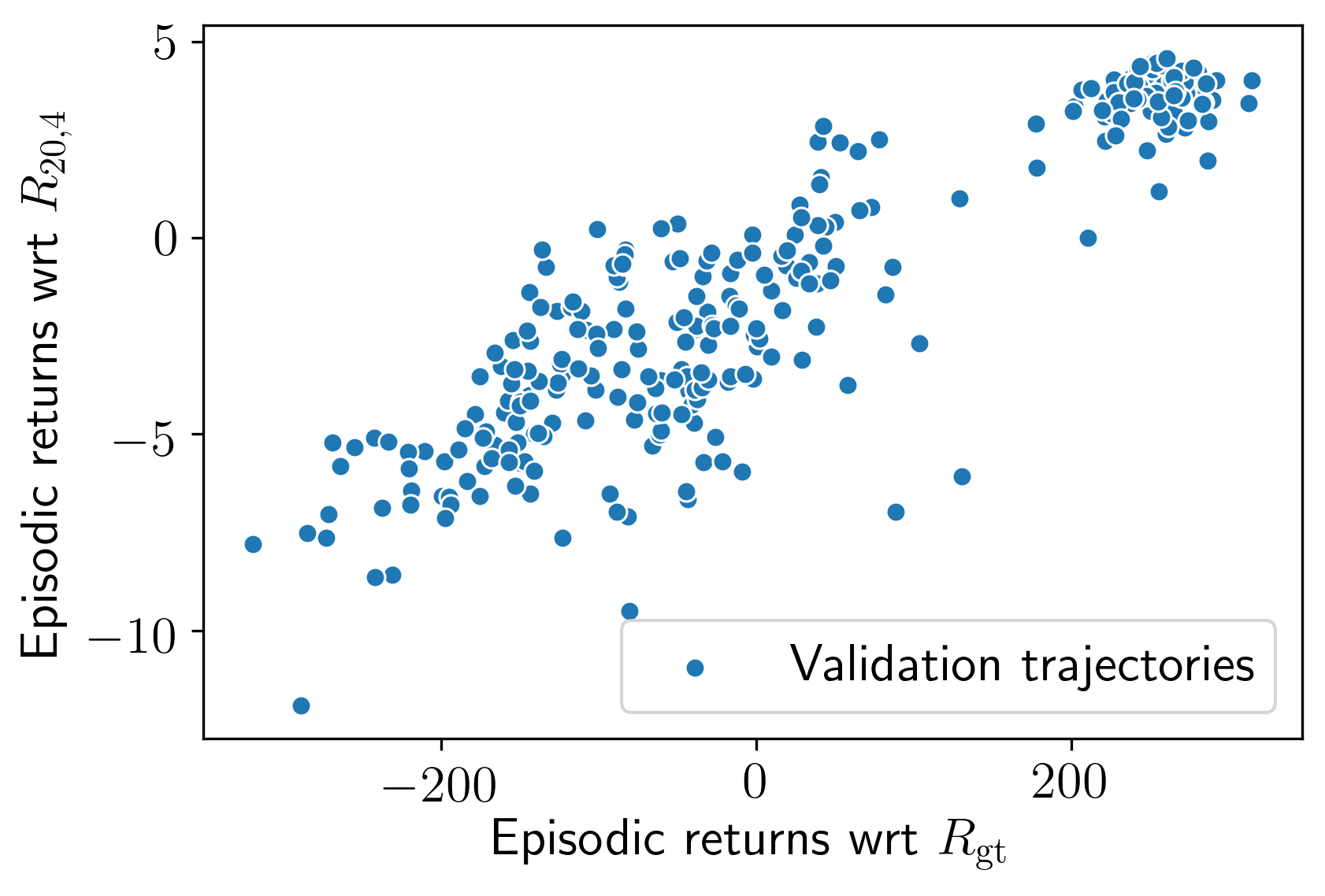}
            \label{fig:162906_validation}
        }
    \end{subfigure}
    \caption{\ref{fig:104400_evaluation_demos} plots ground truth
        episodic returns versus episodic returns with respect to a
        learned reward function $R_{101}$; both were
        evaluated on the demonstrations $\cD_{101}$ that were used for
        learning $R_{101}$, and for trajectories generated by
        policies saved while optimizing a policy for
        $R_{101}$. \ref{fig:200248_evaluation_demos} shows
        the same information for a different reward function
        $R_{100}$ that was trained on different
        demonstrations $\cD_{100}$ (see main
        text). \ref{fig:104400_200248_training_returns} and
        \ref{fig:104400_200248_training_returns_gt} show the
        development of episodic return with respect to $R_{100}$
        resp.\ $R_{101}$ and $R_{\mathrm{gt}}$ during PPO training
        with $R_{100}$ resp.\ $R_{101}$; Figure
        \ref{fig:104400_200248_training_returns_gt} also shows the
        development of episodic returns with respect to
        $R_{\mathrm{gt}}$ when running GAIL with the demonstrations in
        $\cD_{101}$.  \ref{fig:162815_validation} and
        \ref{fig:162906_validation} show evaluations of two trained
        reward functions $R_{50, 4}$ and $R_{20, 4}$ (trained with 50
        resp.\ 20 pairs and 4 fixed points on $\cD_{125}$) on a large
        validation set of trajectories.  In all plots, the learned
        reward functions are on a different scale than
        $R_{\mathrm{gt}}$, as we normalized targets during
        training and did not invert this afterwards, as we found that
        PPO was insensitive to scale changes.}
    \label{fig:episodic_rewards}
\end{figure*}

We tested our approach in OpenAI Gym's LunarLander environment, which
combines the advantages of being fast to train with algorithms like
PPO \cite{schulman_proximal_2017}, and having a sufficiently
interesting ground truth reward function $R_{\mathrm{gt}}$: The agent
receives an immediate small reward or penalty depending on its
velocity, distance to the landing region, angle and use of the
engines, but there is also a long term component in the form of a
bonus of 100 points that the agent can get only once per episode for
landing and turning the engines off. 

We worked with various sets of demonstrations $\cD$ created by
sampling trajectories from a pool of policies of different
qualities. To obtain these, we ran PPO on the ground truth reward
function and saved an intermediate policy after every epoch. To obtain
the optimality profiles $P_\mathrm{tgt}$ required by our algorithm, we
computed the ground truth returns for the trajectories in the
respective augmented set $\widetilde \cD$ \eqref{eq:augmented_trajs}
and produced a histogram (usually with 20 to 50 bins). The
trajectories for which we provided ordered index pairs
$\cI_{\mathrm{pw}}$ were sampled randomly, whereas the (few)
trajectories for which we provided ground truth labels
$\cY_{\mathrm{fix}}$ in addition were selected by hand (we took
trajectories on which $R_\mathrm{gt}^{(\gamma)}$ was close to the
minimum or maximum value present in the respective set $\cD$). We
typically used between 10 and 50 pairs and less than 10 fixed points.

The neural network we used to model the reward function was a simple
MLP with input dimension 8 (reflecting that $R_\mathrm{gt}$ is a
function of 8 features), a single hidden layer of dimension 16 and
ReLU activations. We optimized the reward function with Adam or
RMSprop. For finding a good fit, it usually helped to slowly decrease
the learning rate over time, and also to decrease the entropy constant
in the regularized optimal transport objective
\eqref{eq:ot_objective}. We used a CPU and sometimes an NVIDIA Titan X
GPU for training.

\subsection{Evaluation of fitted reward functions}
\label{sec:eval-fitt-reward}
We evaluated the quality of our learned reward functions in two
different ways: First, we used PPO to train policies optimizing them
and then assessed their performance with respect to $R_{\mathrm{gt}}$.
The results of that were varied: In some cases, we managed to train
policies that were near-optimal at the end of training, meaning that
they achieved an average episodic ground truth return $> 200$, at
which point the environment is viewed as solved, and outperformed
(slightly) the performance of the best demonstrations used for fitting
the reward function, see Figure \ref{fig:104400_evaluation_demos}.

\begin{figure*}[h]
    \centering
    \begin{subfigure}[]{
            \includegraphics[width=0.31\linewidth]{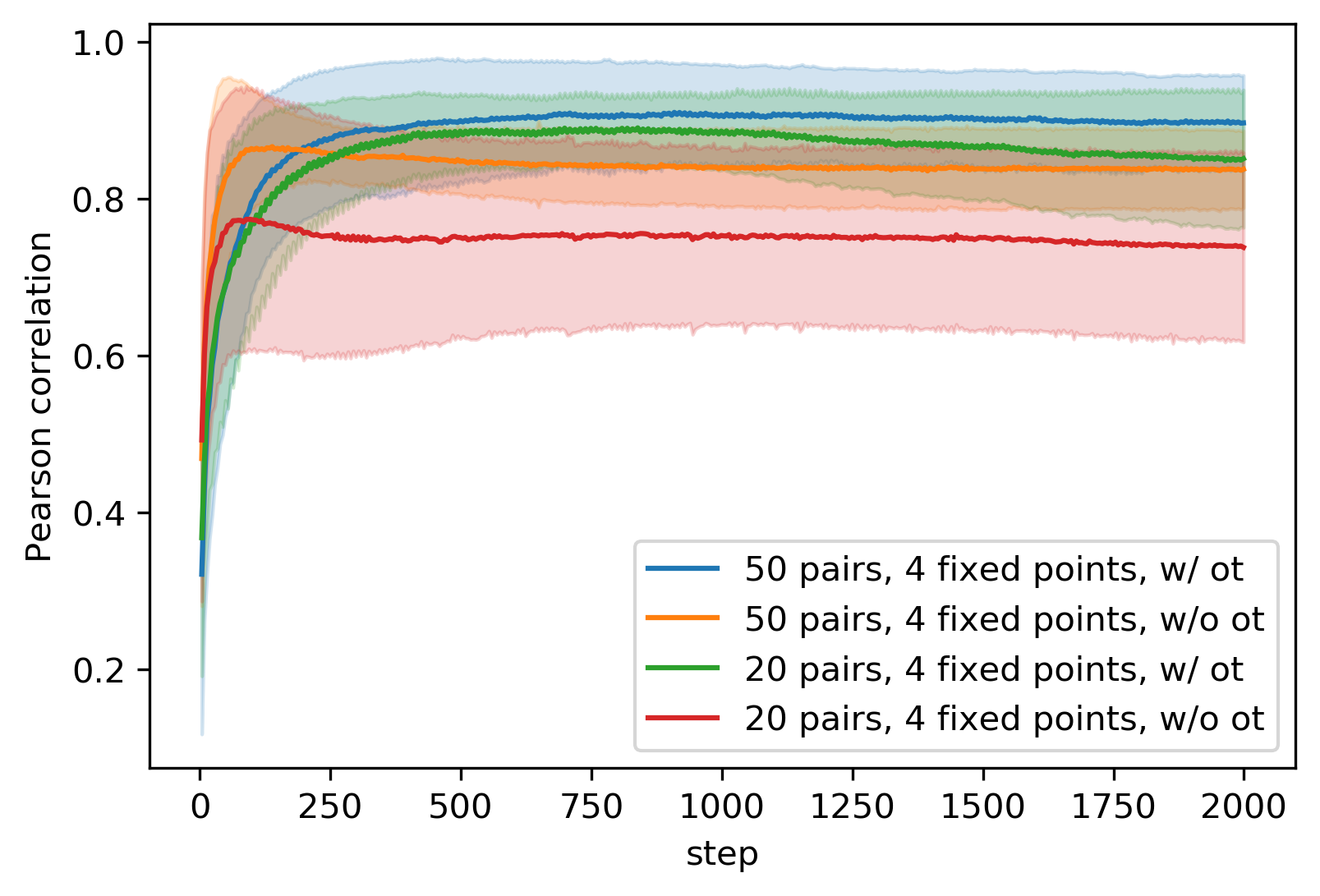}
            \label{fig:ot_vs_pairs__pearson_corr}
        }
    \end{subfigure}
    \begin{subfigure}[]{
            \includegraphics[width=0.31\linewidth]{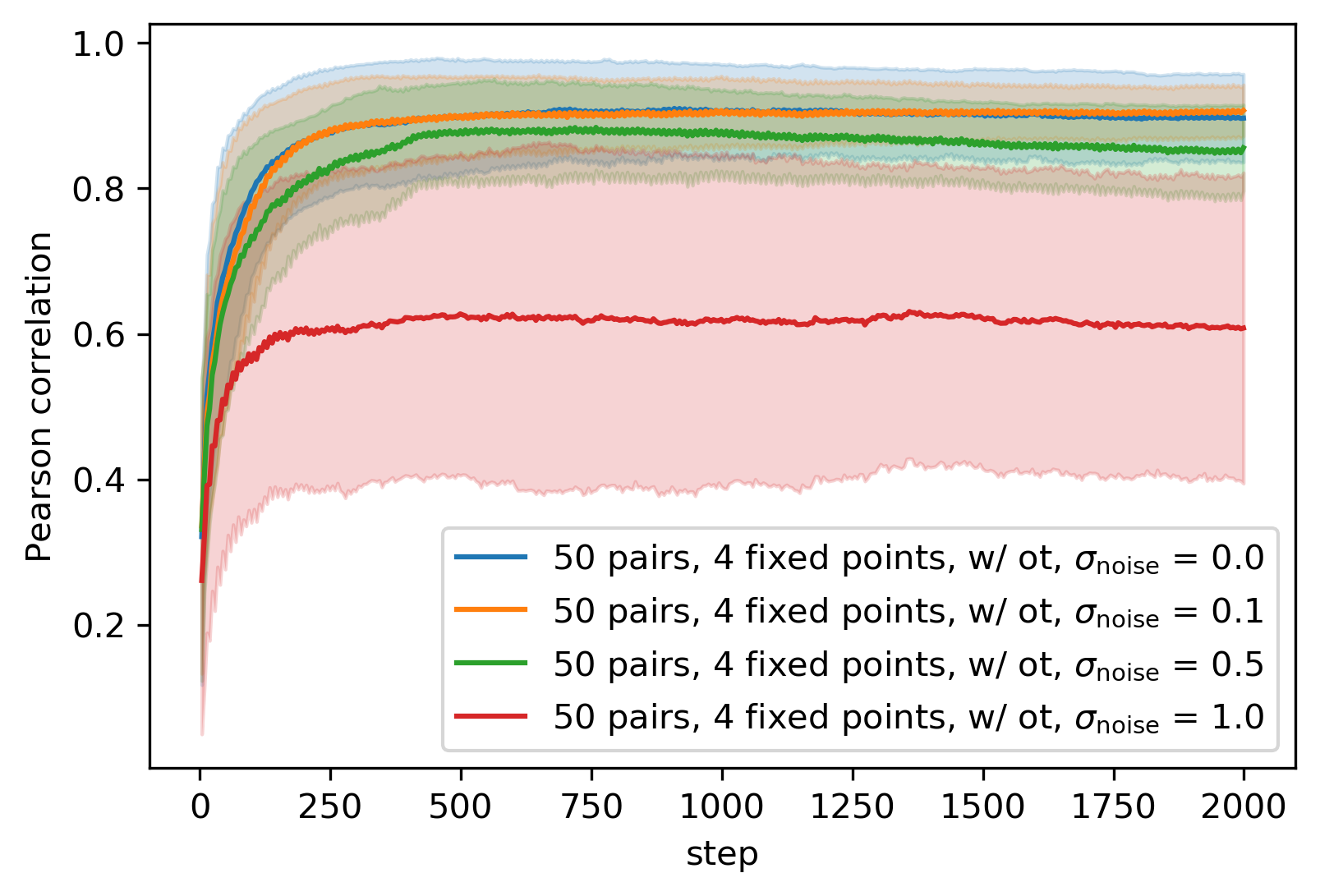}
            \label{fig:noise__pearson_corr}
        }
    \end{subfigure}
    \begin{subfigure}[]{
            \includegraphics[width=0.31\linewidth]{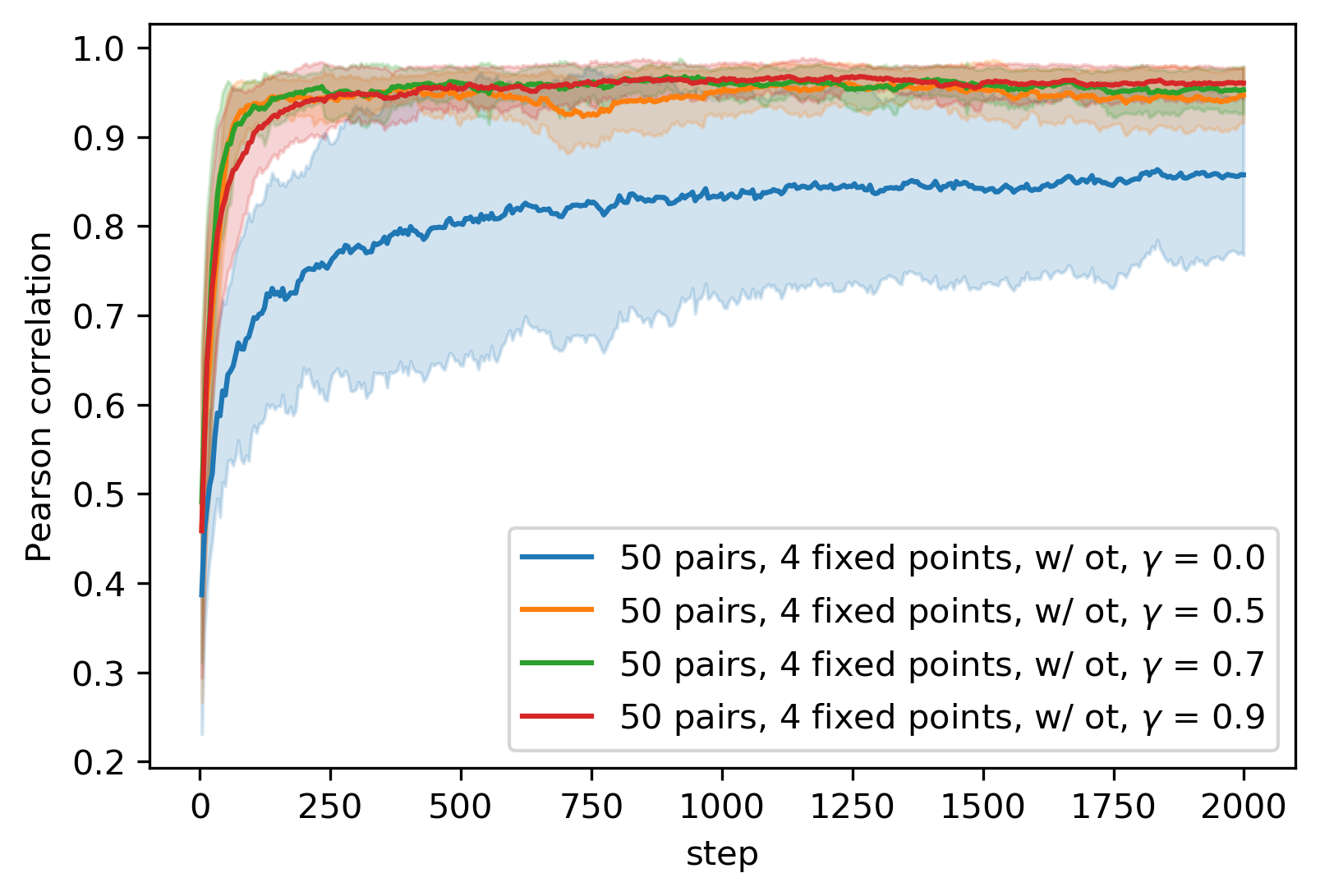}
            \label{fig:gamma__pearson_corr}
        }
    \end{subfigure}
    \caption{\ref{fig:ot_vs_pairs__pearson_corr} plots the Pearson
        correlation between then ground truth and the reward learned
        when our algorithm was given access to 20 resp.\ 50 pairs in addition to an optimality
        profile (``w/ ot''), resp.\ without being given access to an
        optimality profile (``w/o ot''). In all experiments, 4 points
        with ground truth information on the discounted reward were
        provided. 
        \ref{fig:noise__pearson_corr} plots the Pearson correlation
        between the ground truth learned reward and the learned reward for
        various levels of noise applied to the optimality profile
        provided to our algorithm.
        \ref{fig:gamma__pearson_corr} plots the Pearson correlation
        between ground truth and learned reward for various values of the
        discount factor $\gamma \in \{0, 0.5, 0.7, 0.9\}$ 
        used for computing the optimality profile.
        All experiments reported in
        \ref{fig:ot_vs_pairs__pearson_corr},
        \ref{fig:noise__pearson_corr}, \ref{fig:gamma__pearson_corr}
        were run with ten different choices of random seed; the plots
        show means and standard deviations.}
\end{figure*}

Depending on the sets of demonstrations we used, the agent sometimes
ended up with policies whose strategy was (i) to drop to the ground
without using the engines, (ii) to fly out of the screen, or (iii)
approach the landing region and levitate over it. Although suboptimal,
these types of policies have ground truth performance much higher than
random policies. In general, sets of demonstrations with a high
heterogeneity regarding their performance worked better than
demonstrations containing e.g.\ only a few near-optimal
trajectories. This is not surprising: If one starts policy training
with a random policy, it is clearly important to have an estimate of
the reward function which is accurate in regions of the feature space
where near-optimal policies do not spend much time.

One set of demonstrations, $\cD_{100}$ (20000 time steps in total),
contained trajectories collected by running 100 policies of low to
medium performance. Figure \ref{fig:200248_evaluation_demos} shows an
evaluation of the resulting reward function $R_{100}$ both on the
trajectories in $\cD_{100}$ and on trajectories generated by policies
saved while optimizing a policy for $R_{100}$. Note that
while some of the latter achieved higher ground truth performance than
the best demonstration, their performance with respect to
$R_{100}$ is mediocre. In the end, the agent learned to levitate over
the landing region without actually landing. See videos in the
supplementary material.

Another set of demonstrations, $\cD_{101}$ (21000 time steps in total),
was obtained from $\cD_{100}$ by adding three trajectories generated
by a near-optimal policy. Figure \ref{fig:104400_evaluation_demos}
shows that the resulting reward function $R_{101}$ was much better
correlated with the ground truth reward function than $R_{100}$ and
that the policy learned in the end was near-optimal, meaning that the
agent managed to land and turn the engines off (and slightly
outperformed the best demonstration).

Second, we evaluated our learned reward functions on trajectories
sampled from a large set of policies of heterogeneous quality. These
policies were produced by running PPO with various hyperparameter
configurations to encourage diversity, and saving the intermediate
policies after every epoch. Every trajectory in this set corresponded
to one full episode that ended as soon as the agent either reached a
terminal state (landing and coming to rest, or crash) or completed
1000 time steps. Some results of that for reward functions trained on
a large set $\cD_{125}$ (53000 time steps in total) obtained from
$\cD_{100}$ by adding trajectories sampled from 25 good to optimal
policies are shown in Figures \ref{fig:162815_validation} and
\ref{fig:162906_validation}.

\subsection{Comparison with other methods}
\label{sec:comp-with-other}
Our experiments show that when we do not optimize the optimal
transport loss $\cL_{\mathrm{ot}}$ but only the pairwise loss
$\cL_{\mathrm{pw}}$, the ability of our algorithm to to find a good
reward function reduces significantly, unless the number of pairs and
fixed points is increased accordingly, see Figure
\ref{fig:ot_vs_pairs__pearson_corr}. As our method reduces essentially
to the TREX method proposed by \citet{brown_trex_2019} in this case,
this indicates that our method can work with significantly fewer
pairwise comparisons than TREX provided that an optimality profile is
available. We also compared with GAIL \cite{ho2016generative}, which
was not able to find good policies when provided with the
heterogeneous sets of trajectories that we provided to our algorithm,
see Figure \ref{fig:104400_200248_training_returns_gt}.

\subsection{Robustness to noise}
\label{sec:robustness}
We also evaluated the robustness of our method with respect to noise
in the optimality profile $P_\mathrm{tgt}$ that was provided to our
algorithm. 
To achieve that, we multiplied the cumulative discounted reward values
from which the optimality profile was computed with a noise factor
sampled from $\cN(1, \sigma_{\mathrm{noise}})$ for
$\sigma_{\mathrm{noise}} \in \{0.1, 0.5, 1\}$ and provided the
resulting histogram to the algorithm. The correlation plot in Figure
\ref{fig:noise__pearson_corr} shows that, while noise clearly affects the
ability of the algorithm to match the ground truth closely, small
amounts of noise seem to be tolerable.

\subsection{Effect of varying the discount factor $\gamma$}
\label{sec:effects_parameters}
We observed that higher values of $\gamma$ tend to lead to better fits
of $R_\mathrm{gt}$ by the learned reward function, as shown by Figure
\ref{fig:gamma__pearson_corr} which plots the development $L_1$-distance
between the two functions during training for various values of
$\gamma$. This may be counter-intuitive, as the supervision that the
algorithm gets is computed using $R^{(\gamma)}_{\mathrm{gt}}$, which
in some sense approximates $R_{\mathrm{gt}}$ as $\gamma$ gets
smaller. However, it seems that the requirement of matching
$P_\mathrm{tgt} = P_{R_{\mathrm{gt}}}^{(\gamma)}$ actually imposes a stronger constraint on
the reward function to be learned as $\gamma$ gets larger.

%% file: conclusions.tex
\section{Discussion and outlook}
\label{sec:conclusions}
We presented an IRL algorithm that fits a reward function to
demonstrations by minimizing the Wasserstein distance to a given
optimality profile, that is, a distribution over (cumulative
discounted future) rewards which we view as an assessment of the
degree of optimality of the demonstrations. To the best of our
knowledge, this is the first IRL algorithm that works with this kind
of supervision signal. In future work, we will investigate ways of
effectively eliciting such optimality profiles from human experts. We
also plan to study the objective we propose from an optimization
perspective, and to evaluate the performance of our method in
high-dimensional settings.

%% file: appendix.tex
\section{Additional explanations regarding state and trajectory distributions}
\label{app:setting}

\subsection{Distributions on state and trajectory spaces}
\label{sec:distr-traj-stat}

In the main paper, we claimed that one could interpret the
distribution $\rho_{\cT_1}$ on $\cT \times [0, T]$ resp.\
$\cT_1 = \{(\bs, t) \in \cT \times [0, T] ~|~ t \leq \ell(\bs) - 1\}$
by saying that $\rho_{\cT_1}(\bs, t)$ is the probability of obtaining
$(\bs, t)$ if one first samples a trajectory from $\cT$ according to a
version of $P_\mathrm{traj}'$ in which probabilities are rescaled
proportionally to trajectory length, and then a time step $t$
uniformly at random from $[0, \dots \ell(\bs) - 1]$. To see that this
is indeed the case, note that the rescaled version of
$P_\mathrm{traj}'$ is given by
\begin{equation}
    \label{eq:1}
    \widetilde{P}_{\mathrm{traj}}'(\bs) =
    \frac{\ell(\bs) P'_{\mathrm{traj}}(\bs)}{\sum_{\bs' \in
            \cT} \ell(\bs') P'_{\mathrm{traj}}(\bs')}.
\end{equation}
Now if we sample $\bs \sim \widetilde{P}_{\mathrm{traj}}'$ and then
$t$ uniformly at random from $[0, \dots, \ell(\bs) - 1]$, the
probability of the resulting $(\bs, t)$ is
\begin{equation}
    \label{eq:2}
    \widetilde P_{\mathrm{traj}}'(\bs) \cdot \frac{1}{\ell(\bs)} =
    \frac{P'_{\mathrm{traj}}(\bs)}{\sum_{\bs' \in \cT} \ell(\bs')
        P_\mathrm{traj}'(\bs')},
\end{equation}
which is the expression for $\rho_{\mathrm{\cT_1}}(\bs, t)$ given in
equation \eqref{eq:traj_timestep_distr} in the main paper.

\begin{figure}[h]
    \centering
    \includegraphics[scale=0.75]{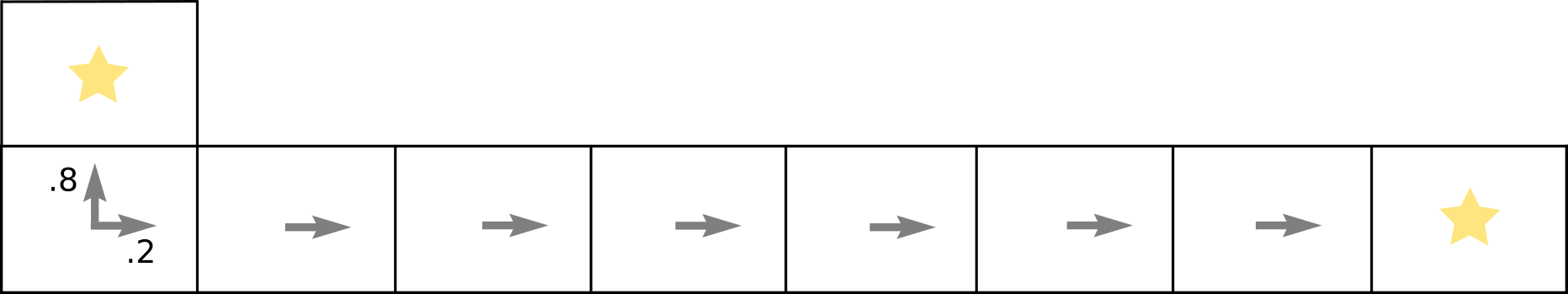}
    \label{fig:toy_gridworld}
    \caption{A gridworld to illustrate the difference between the
        non-rescaled trajectory distribution $P_\mathrm{traj}'$ and
        its rescaled version $\widetilde{P}_\mathrm{traj}$. We assume
        an initial state distribution whose mass is concentrated on
        the state in the lower left corner, so that all trajectories
        start there. The two states marked with a star are terminal
        states. The arrows represent a policy $\pi$ which with
        probability 0.2 moves upwards from the start state and thus
        ends up in a terminal state immediately, and with probability
        0.2 moves to the right from the start state and then continues
        in that direction until it reaches the terminal state.}
\end{figure}

To get some intuition for why we consider the \emph{rescaled}
distribution over trajectories, consider the grid world depicted in
Figure \ref{fig:toy_gridworld} below and the policy $\pi$ described
there. The only two possible trajectories in this setting are a
``vertical'' trajectory $\bs_\mathrm{v}$ of length 2, and a
``horizonal'' trajectory $\bs_\mathrm{h}$ of length 8. Note that the
non-rescaled trajectory distribution has
$P_{\mathrm{traj}}'(\bs_\mathrm{v}) = 0.8$ and
$P_{\mathrm{traj}}'(\bs_\mathrm{h}) = 0.2$, whereas the rescaled
trajectory distribution has
$\widetilde{P}_{\mathrm{traj}}'(\bs_\mathrm{v}) = 0.5$ and
$\widetilde{P}_{\mathrm{traj}}'(\bs_\mathrm{h}) = 0.5$, reflecting the
fact that, in expectation, $\pi$ spends the same amount of time in the
vertical as in the horizontal trajectory. We believe that in many
situations in which for example human experts are asked to judge
behavior it is natural to consider distributions over trajecories that
take trajectory length into account, which is why we chose to work
with the rescaled trajectory distribution in this paper. However, the
method proposed in the paper can also be adapted to other choices.

\subsection{Distributions induced by the reward function}
We also claimed that the reward distribution $P_R$ could be recovered
as the special case of the distribution of cumulative discounted
rewards $P_R^{(\gamma)}$ for $\gamma = 0$, i.e., $P_R^{(0)} = P_R$. To
see that, recall that these distributions are given by
\begin{equation}
    P_R^{(0)} = R^{(0)}_* \rho_\cT = (R^{(0)})_*((\Pi_\cT)_*
    \rho_{\cT_1}) = (R^{(0)} \circ \Pi_\cT)_*
    \rho_{\cT_1}
    \label{eq:disc_reward_distr_0_app}
\end{equation}
resp.\
\begin{equation}
    \label{eq:reward_distr_app}
    P_R = R_* \rho_\cS = R_* ((\Pi_\cS)_* \rho_{\cT_1}) = (R \circ
    \Pi_\cS)_* \rho_{\cT_1},
\end{equation}
where the maps are those introduced in Section
\ref{sec:setting}. Note that these maps fit in a diagram
\begin{equation}
    \begin{aligned}
        \xymatrix{
            \cT_1 \ar[r]^{\Pi_\cT}\ar[d]_{\Pi_\cS} & \cT \ar[d]^{R^{(0)}} \\
            \cS \ar[r]^{R}                         & \R 
        }
    \end{aligned}
\end{equation}
which commutes, meaning that
\begin{equation}
    \label{eq:equality-of-compositions}
    R^{(0)} \circ \Pi_{\cT} = R \circ \Pi_{\cS}
\end{equation}
Indeed, for
$(\bs, t) \in \cT_1$, we have
\begin{equation}
    \label{eq:3}
    R^{(0)} (\Pi_\cT(\bs, t)) = R^{(0)}(s_t, \dots, s_{\ell(\bs) -
        1})= R(s_t) = R (\Pi_{\cS}(\bs, t)).
\end{equation}
\eqref{eq:equality-of-compositions} together with
\eqref{eq:disc_reward_distr_0_app} and \eqref{eq:reward_distr_app}
implies the claimed equality of distributions $P_R^{(0)} = P_R$.

%% file: main.bbl
\begin{thebibliography}{27}
\providecommand{\natexlab}[1]{#1}
\providecommand{\url}[1]{\texttt{#1}}
\expandafter\ifx\csname urlstyle\endcsname\relax
  \providecommand{\doi}[1]{doi: #1}\else
  \providecommand{\doi}{doi: \begingroup \urlstyle{rm}\Url}\fi

\bibitem[Abbeel \& Ng(2004)Abbeel and Ng]{abbeel2004apprenticeship}
Abbeel, P. and Ng, A.~Y.
\newblock Apprenticeship learning via inverse reinforcement learning.
\newblock In \emph{ICML}, 2004.

\bibitem[Akrour et~al.(2011)Akrour, Schoenauer, and
  Sebag]{akrour2011preference}
Akrour, R., Schoenauer, M., and Sebag, M.
\newblock Preference-based policy learning.
\newblock In \emph{Joint European Conference on Machine Learning and Knowledge
  Discovery in Databases}, pp.\  12--27. Springer, 2011.

\bibitem[Arjovsky et~al.(2017)Arjovsky, Chintala, and
  Bottou]{arjovsky_wasserstein_2017}
Arjovsky, M., Chintala, S., and Bottou, L.
\newblock Wasserstein {GAN}.
\newblock \emph{arXiv:1701.07875 [cs, stat]}, January 2017.
\newblock URL \url{http://arxiv.org/abs/1701.07875}.
\newblock arXiv: 1701.07875.

\bibitem[Bellemare et~al.(2017)Bellemare, Dabney, and
  Munos]{bellemare2017distributional}
Bellemare, M.~G., Dabney, W., and Munos, R.
\newblock A distributional perspective on reinforcement learning.
\newblock In \emph{Proceedings of the 34th International Conference on Machine
  Learning-Volume 70}, pp.\  449--458. JMLR. org, 2017.

\bibitem[Bradley \& Terry(1952)Bradley and Terry]{bradley_rank_1952}
Bradley, R.~A. and Terry, M.~E.
\newblock Rank {Analysis} of {Incomplete} {Block} {Designs}: {I}. {The}
  {Method} of {Paired} {Comparisons}.
\newblock \emph{Biometrika}, 39\penalty0 (3/4):\penalty0 324--345, 1952.
\newblock ISSN 0006-3444.
\newblock \doi{10.2307/2334029}.
\newblock URL \url{https://www.jstor.org/stable/2334029}.

\bibitem[Brown et~al.(2019{\natexlab{a}})Brown, Goo, Nagarajan, and
  Niekum]{brown_trex_2019}
Brown, D., Goo, W., Nagarajan, P., and Niekum, S.
\newblock Extrapolating {Beyond} {Suboptimal} {Demonstrations} via {Inverse}
  {Reinforcement} {Learning} from {Observations}.
\newblock In \emph{International {Conference} on {Machine} {Learning}}, pp.\
  783--792, May 2019{\natexlab{a}}.
\newblock URL \url{http://proceedings.mlr.press/v97/brown19a.html}.

\bibitem[Brown et~al.(2019{\natexlab{b}})Brown, Goo, and
  Niekum]{brown_drex_2019}
Brown, D.~S., Goo, W., and Niekum, S.
\newblock Better-than-{Demonstrator} {Imitation} {Learning} via
  {Automatically}-{Ranked} {Demonstrations}.
\newblock \emph{arXiv:1907.03976 [cs, stat]}, October 2019{\natexlab{b}}.
\newblock URL \url{http://arxiv.org/abs/1907.03976}.
\newblock arXiv: 1907.03976.

\bibitem[Christiano et~al.(2017)Christiano, Leike, Brown, Martic, Legg, and
  Amodei]{christiano_deep_2017}
Christiano, P., Leike, J., Brown, T.~B., Martic, M., Legg, S., and Amodei, D.
\newblock Deep reinforcement learning from human preferences.
\newblock \emph{arXiv:1706.03741 [cs, stat]}, June 2017.
\newblock URL \url{http://arxiv.org/abs/1706.03741}.
\newblock arXiv: 1706.03741.

\bibitem[Cuturi(2013)]{cuturi_sinkhorn_2013}
Cuturi, M.
\newblock Sinkhorn {Distances}: {Lightspeed} {Computation} of {Optimal}
  {Transport}.
\newblock In \emph{Advances in {Neural} {Information} {Processing} {Systems}
  26}, pp.\  2292--2300. Curran Associates, Inc., 2013.

\bibitem[Finn et~al.(2016)Finn, Levine, and Abbeel]{finn2016guided}
Finn, C., Levine, S., and Abbeel, P.
\newblock Guided cost learning: Deep inverse optimal control via policy
  optimization.
\newblock In \emph{ICML}, pp.\  49--58, 2016.

\bibitem[Flamary \& Courty(2017)Flamary and Courty]{flamary2017pot}
Flamary, R. and Courty, N.
\newblock Pot python optimal transport library, 2017.
\newblock URL \url{https://github.com/rflamary/POT}.

\bibitem[Fu et~al.(2017)Fu, Luo, and Levine]{fu2017learning}
Fu, J., Luo, K., and Levine, S.
\newblock Learning robust rewards with adversarial inverse reinforcement
  learning.
\newblock \emph{arXiv preprint arXiv:1710.11248}, 2017.

\bibitem[Ho \& Ermon(2016)Ho and Ermon]{ho2016generative}
Ho, J. and Ermon, S.
\newblock Generative adversarial imitation learning.
\newblock In \emph{NIPS}, 2016.

\bibitem[Ibarz et~al.(2018)Ibarz, Leike, Pohlen, Irving, Legg, and
  Amodei]{ibarz_reward_2018}
Ibarz, B., Leike, J., Pohlen, T., Irving, G., Legg, S., and Amodei, D.
\newblock Reward learning from human preferences and demonstrations in {Atari}.
\newblock \emph{arXiv:1811.06521 [cs, stat]}, November 2018.
\newblock URL \url{http://arxiv.org/abs/1811.06521}.
\newblock arXiv: 1811.06521.

\bibitem[Levine(2018)]{levine_reinforcement_2018}
Levine, S.
\newblock Reinforcement {Learning} and {Control} as {Probabilistic}
  {Inference}: {Tutorial} and {Review}.
\newblock \emph{arXiv:1805.00909 [cs, stat]}, May 2018.
\newblock URL \url{http://arxiv.org/abs/1805.00909}.
\newblock arXiv: 1805.00909.

\bibitem[Luce(1959)]{luce_individual_1959}
Luce, R.~D.
\newblock \emph{Individual choice behavior}.
\newblock John Wiley, Oxford, England, 1959.

\bibitem[Ng \& Russell(2000)Ng and Russell]{ng2000algorithms}
Ng, A.~Y. and Russell, S.~J.
\newblock Algorithms for inverse reinforcement learning.
\newblock In \emph{ICML}, 2000.

\bibitem[Peyré \& Cuturi(2018)Peyré and Cuturi]{peyre_computational_2018}
Peyré, G. and Cuturi, M.
\newblock Computational {Optimal} {Transport}.
\newblock \emph{arXiv:1803.00567 [stat]}, March 2018.
\newblock URL \url{http://arxiv.org/abs/1803.00567}.
\newblock arXiv: 1803.00567.

\bibitem[Schulman et~al.(2017)Schulman, Wolski, Dhariwal, Radford, and
  Klimov]{schulman_proximal_2017}
Schulman, J., Wolski, F., Dhariwal, P., Radford, A., and Klimov, O.
\newblock Proximal {Policy} {Optimization} {Algorithms}.
\newblock \emph{arXiv:1707.06347 [cs]}, July 2017.
\newblock URL \url{http://arxiv.org/abs/1707.06347}.
\newblock arXiv: 1707.06347.

\bibitem[Shiarlis et~al.(2016)Shiarlis, Messias, and
  Whiteson]{shiarlis_inverse_2016}
Shiarlis, K., Messias, J., and Whiteson, S.
\newblock Inverse {Reinforcement} {Learning} from {Failure}.
\newblock In \emph{Proceedings of the 2016 {International} {Conference} on
  {Autonomous} {Agents} \& {Multiagent} {Systems}}, {AAMAS} '16, pp.\
  1060--1068, 2016.
\newblock ISBN 978-1-4503-4239-1.
\newblock URL \url{http://dl.acm.org/citation.cfm?id=2936924.2937079}.

\bibitem[Warnell et~al.(2018)Warnell, Waytowich, Lawhern, and
  Stone]{warnell2018deep}
Warnell, G., Waytowich, N., Lawhern, V., and Stone, P.
\newblock Deep tamer: Interactive agent shaping in high-dimensional state
  spaces.
\newblock In \emph{Thirty-Second AAAI Conference on Artificial Intelligence},
  2018.

\bibitem[Wilson et~al.(2012)Wilson, Fern, and Tadepalli]{wilson2012bayesian}
Wilson, A., Fern, A., and Tadepalli, P.
\newblock A bayesian approach for policy learning from trajectory preference
  queries.
\newblock In \emph{Advances in neural information processing systems}, pp.\
  1133--1141, 2012.

\bibitem[Wirth et~al.(2017)Wirth, Akrour, Neumann, and
  F{\"u}rnkranz]{wirth2017survey}
Wirth, C., Akrour, R., Neumann, G., and F{\"u}rnkranz, J.
\newblock A survey of preference-based reinforcement learning methods.
\newblock \emph{The Journal of Machine Learning Research}, 18\penalty0
  (1):\penalty0 4945--4990, 2017.

\bibitem[Wulfmeier et~al.(2015)Wulfmeier, Ondruska, and
  Posner]{wulfmeier2015maximum}
Wulfmeier, M., Ondruska, P., and Posner, I.
\newblock Maximum entropy deep inverse reinforcement learning.
\newblock \emph{arXiv:1507.04888}, 2015.

\bibitem[Xiao et~al.(2019)Xiao, Herman, Wagner, Ziesche, Etesami, and
  Linh]{xiao2019wasserstein}
Xiao, H., Herman, M., Wagner, J., Ziesche, S., Etesami, J., and Linh, T.~H.
\newblock Wasserstein adversarial imitation learning.
\newblock \emph{arXiv preprint arXiv:1906.08113}, 2019.

\bibitem[Ziebart et~al.(2008)Ziebart, Maas, Bagnell, and
  Dey]{ziebart2008maximum}
Ziebart, B.~D., Maas, A.~L., Bagnell, J.~A., and Dey, A.~K.
\newblock Maximum entropy inverse reinforcement learning.
\newblock In \emph{AAAI}, 2008.

\bibitem[Ziebart et~al.(2013)Ziebart, Bagnell, and Dey]{ziebart2013principle}
Ziebart, B.~D., Bagnell, J.~A., and Dey, A.~K.
\newblock The principle of maximum causal entropy for estimating interacting
  processes.
\newblock \emph{IEEE Transactions on Information Theory}, 59\penalty0
  (4):\penalty0 1966--1980, 2013.

\end{thebibliography}
